\documentclass{optica-article}

\journal{opticajournal}
\articletype{Research Article}

\usepackage{pifont}
\usepackage{url}
\usepackage{float}

\begin{document}

\title{Local--Global Joint Modeling for Point Cloud Super-Resolution Guided by Multi-Scale Spatial Priors}

\author{WANG YANG,\authormark{1} WENYI GE,\authormark{1,*}}

\address{\authormark{1}College of Computer Science, Chengdu University of Information Technology, Chengdu 610225, China}

\email{\authormark{*}gewenyi15@cuit.edu.cn}

\begin{abstract*}

Range-image-based point cloud super-resolution provides an effective way to reconstruct sparse outdoor point clouds by projecting low-resolution LiDAR scans into a regular 2D representation. However, current methods still face three challenges. First, range images exhibit pronounced vertical nonuniformity, with sparse observations in distant regions and denser observations in nearby regions. Some cross-scale fusion mechanisms do not account for the varying reliability of features at different vertical positions, allowing invalid values and noise from distant sparse regions to propagate into the decoder. Second, a single convolutional or sequence-modeling architecture often struggles to preserve both local high-frequency geometric details and global long-range dependencies, resulting in over-smoothed reconstructions. Third, smooth transition depths near foreground-background boundaries can produce spike artifacts when the 2D range image is back-projected into 3D. To address these issues, we propose a point cloud super-resolution method based on multi-scale spatial priors and joint local-global modeling. The design comprises a spatially gated fusion module, a multi-granularity local-global cooperative perception module, and 2D median-residual denoising. Experiments on KITTI-360 and nuScenes demonstrate effective reconstruction, with IoU scores of 0.4703 and 0.3593, respectively, supporting the feasibility of the proposed method for range-image super-resolution.

\end{abstract*}

\section{Introduction}

LiDAR actively captures the positions, distances, and geometric contours of objects in 3D scenes, providing reliable spatial data for environmental perception. It is therefore widely used in autonomous driving, robot navigation, scene reconstruction, and localization\cite{ref1,ref2,ref3,ref4,ref5,ref6}. However, due to the scanning mechanism and operating principles of LiDAR, point density decreases with increasing range. Distant regions are consequently sparse and often lack essential geometric details, making it difficult to meet the completeness and boundary-clarity requirements of high-precision perception. Point cloud super-resolution aims to recover denser representations from low-beam or otherwise sparse inputs. The challenge is to reconstruct missing spatial structures without introducing erroneous smoothing or artifacts around boundaries, depth discontinuities, and distant regions. Existing research has mainly followed two paradigms: processing point clouds directly in 3D space and modeling them through range images.

Methods that process point clouds directly in 3D include point-based, voxel-based, and graph-convolution approaches\cite{ref3,ref7,ref8,ref9,ref10,ref11,ref12,ref13,ref14,ref15}. Point-based methods extract features and increase the number of points directly from unordered 3D coordinates, but local neighborhoods must be constructed explicitly, making it difficult to preserve global structural consistency at long range. Voxel-based methods discretize point clouds into regular 3D grids for feature aggregation, but quantization constrains output detail to the voxel resolution, while the many empty voxels limit the representation of fine geometry. Graph-convolution methods model relationships between points through graph structures; in large outdoor scenes, however, dynamic graph topologies are susceptible to noise and sparse sampling, which can blur boundaries or omit distant structures.

Range-image-based point cloud super-resolution first projects an unstructured point cloud onto a regular 2D range image according to the spherical scanning geometry of the LiDAR\cite{ref1,ref2,ref4,ref5,ref6,ref16,ref17,ref18,ref19,ref20,ref21,ref22}. Features are then extracted on the 2D grid, the resolution is increased, and depth values are predicted. Finally, the high-resolution range image is back-projected into a dense 3D point cloud. By transforming reconstruction of an irregular point set into a structured 2D image-reconstruction problem, this paradigm makes it easier to exploit 2D neighborhood continuity and established image super-resolution architectures. It has consequently become a major approach to outdoor LiDAR point cloud super-resolution.

Despite this structured representation, range-image methods may still face three reconstruction challenges related to the properties of range images. First, their vertical distribution is strongly nonuniform: distant regions often contain fewer valid observations than nearby regions, producing a sparse-upper, dense-lower pattern. When cross-layer fusion does not explicitly model feature reliability at different vertical positions, invalid values and noise from distant sparse regions can persist across multiple scales and hinder structural recovery. Second, feature extractors based only on local convolutions or sequence modeling often struggle to preserve both local high-frequency geometry and global long-range dependencies across different data distributions. Purely convolutional networks have limited receptive fields, whereas global modeling without boundary constraints may weaken geometric discontinuities\cite{ref24,ref25,ref26,ref27} and over-smooth the reconstruction, as illustrated in Fig.~1. Third, transition depths near foreground-background boundaries can become floating points or elongated artifacts after back-projection, compromising 3D surface consistency, as shown in Fig.~2.

\begin{figure}[tbp]

\centering

\includegraphics[width=\linewidth]{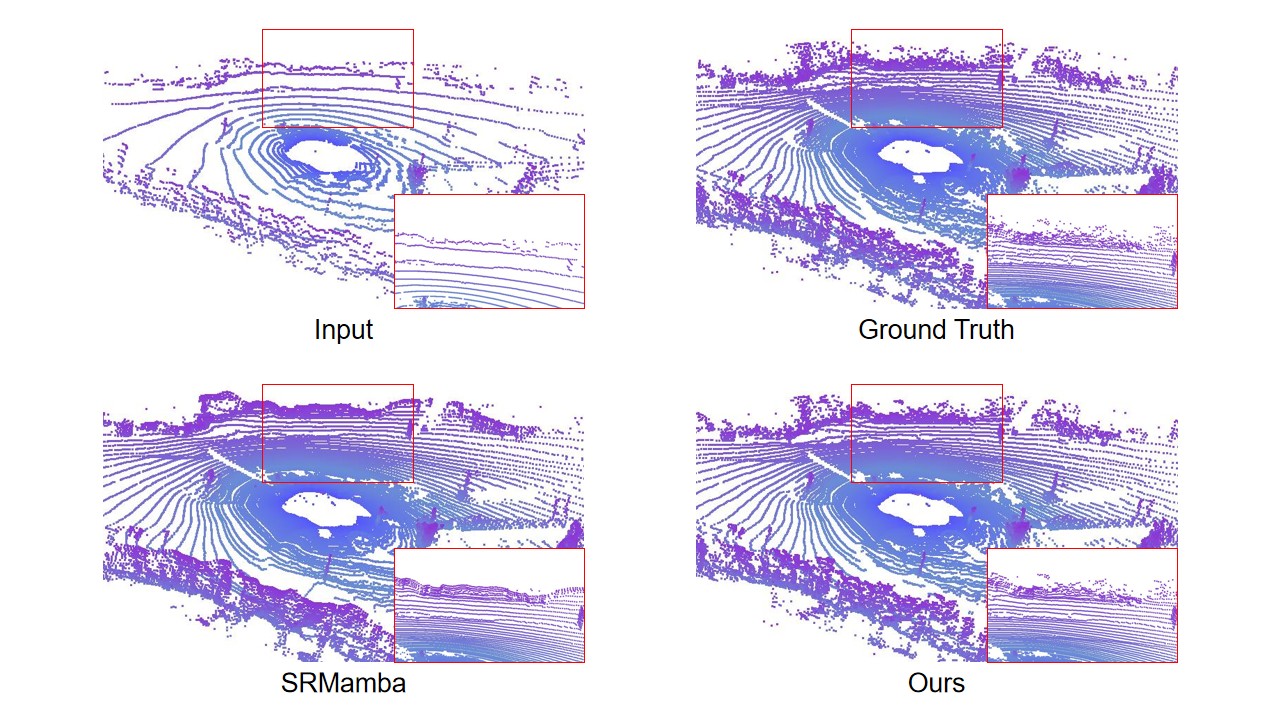}

\caption{Illustration of the point cloud over-smoothing problem}

\label{fig:smoothing}

\end{figure}

\begin{figure}[tbp]

\centering

\includegraphics[width=\linewidth]{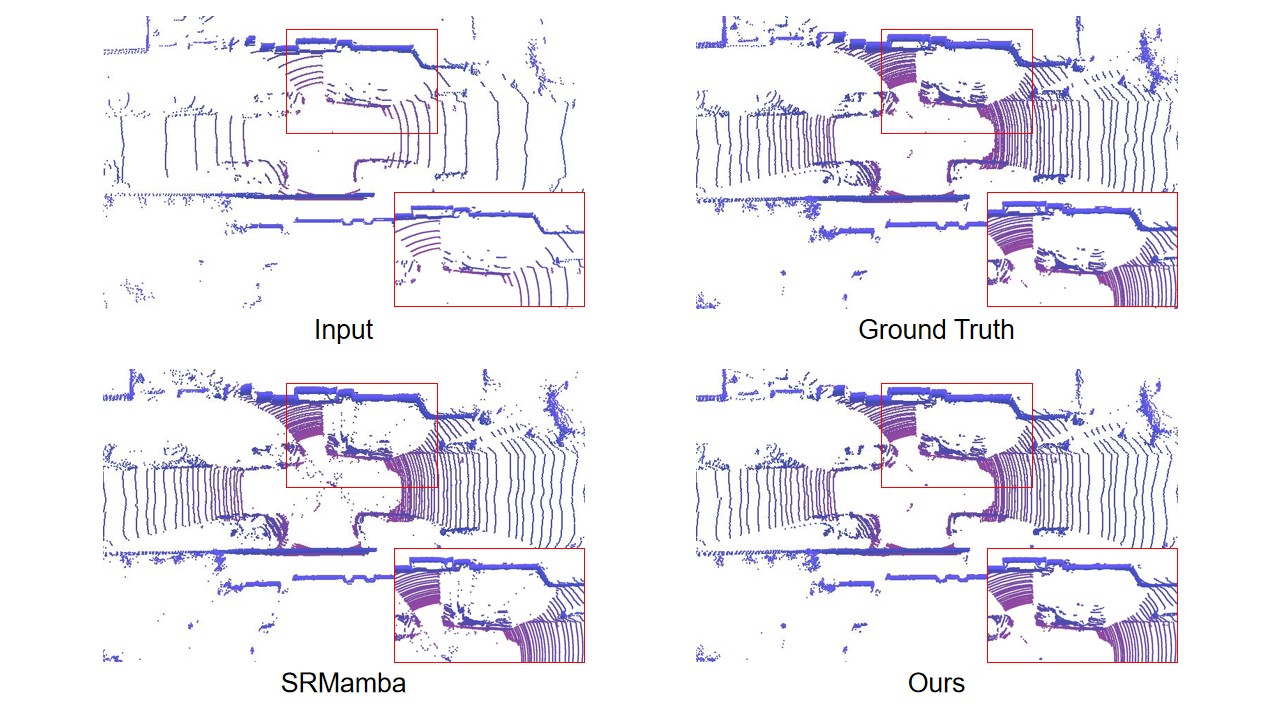}

\caption{Illustration of spike artifacts in the point cloud}

\label{fig:artifacts}

\end{figure}

To address these issues, we propose a point cloud super-resolution method based on multi-scale spatial priors and joint local-global modeling. The method targets three key challenges in range images: sparse distant regions, local boundaries that are prone to smoothing, and spike artifacts introduced during back-projection. It introduces corresponding constraints at the feature-fusion, feature-modeling, and post-processing stages. Specifically, the spatially gated fusion module uses vertical spatial priors to regulate skip-feature transfer and provide targeted compensation in distant sparse regions. The multi-granularity local-global cooperative perception module combines local convolution with selective state-space modeling\cite{ref23,ref24,ref25,ref26,ref28,ref29,ref30,ref31,ref32,ref33}, preserving edge details while establishing long-range dependencies. The 2D median-residual denoising step identifies anomalous residuals using neighborhood continuity in the image domain before back-projection, suppressing spike artifacts at their source. The main contributions are as follows:

(1) We propose SpatialGateFuse, a spatially gated fusion module for the nonuniform, sparse-upper and dense-lower distribution of range images. It introduces vertical spatial priors and content-aware gating to adaptively enhance shallow features in skip connections, mitigating insufficient feature transfer in distant sparse regions.

(2) We propose the multi-granularity local-global cooperative perception module (LG-SSM Block). It complements the local edge-extraction capability of depthwise separable convolutions with the long-range dependency modeling of a selective state-space model, addressing the difficulty of preserving both local geometric detail and global continuity with a single convolutional or sequence-modeling architecture.

(3) We propose a 2D median-residual denoising method. To handle transition depths near foreground-background boundaries, it computes residuals between predicted depths and local median references in the range-image domain, then filters anomalous pixels before back-projection. This suppresses 3D spike artifacts while preserving genuine geometric edges.

\section{Related Work}

\subsection{Point Cloud Super-Resolution}

Point cloud super-resolution reconstructs dense, high-quality 3D point clouds from sparse, low-beam LiDAR scans\cite{ref34,ref35,ref36,ref37,ref38,ref39,ref40,ref41,ref42,ref43,ref44,ref45,ref46,ref47}. Direct 3D methods face inherent challenges arising from irregular representations and topology construction. Point-based methods extract features from unordered coordinates, but require frequent neighborhood searches to construct local topology, making global consistency difficult to maintain. Voxel-based methods discretize point clouds into 3D grids; as resolution increases, voxel counts grow cubically, limiting output accuracy to computationally feasible grid sizes, while empty voxels provide little information about fine geometry. Graph-convolution methods model point relationships with dynamic graphs, but frequent topology updates in large outdoor scenes make long-range information propagation vulnerable to noise and can degrade boundary reconstruction.

Range-image methods project unstructured 3D point clouds onto regular 2D grids using spherical projection\cite{ref1,ref2,ref4,ref5,ref6,ref16,ref17,ref18,ref19,ref20,ref21,ref22}, avoiding the irregular representations and topology-construction challenges of direct 3D processing. They have become a mainstream approach to point cloud super-resolution. Early range-image methods used simple convolutional networks to demonstrate the feasibility of this paradigm. However, several reconstruction challenges remain unresolved.

Some range-image methods adopt the direct channel-concatenation strategy of standard U-Net architectures\cite{ref23}, passing encoder features at different levels to the decoder without assessing their reliability. LiDAR projection and scene distributions make vertical sampling nonuniform: in some fields of view, distant regions in the upper image are sparse while nearby regions in the lower image are denser. Because distant sparse regions contain fewer valid observations, invalid responses make up a larger proportion of their features and are more likely to interfere with structural completion. The limited receptive field of purely convolutional architectures\cite{ref32} can also constrain modeling of panoramic horizontal continuity. Although global attention or sequence modules expand the receptive field, without boundary-aware designs they may smooth high-frequency geometric discontinuities. Implausible depth transitions at foreground-background boundaries can produce floating points or elongated artifacts after back-projection; 3D clustering in post-processing does not directly exploit 2D neighborhood continuity in the range image.

In contrast, this work improves range-image super-resolution at three levels. SpatialGateFuse introduces spatial-prior-guided adaptive skip fusion to address feature transfer under nonuniform distributions. LG-SSM Block couples depthwise separable convolutions with a selective state-space model to preserve local high-frequency detail and global long-range dependencies. Finally, 2D median-residual denoising exploits edge-preserving behavior in the image domain to remove back-projection artifacts.

\subsection{State-Space Models in Vision}

State-space models (SSMs) have shown strong potential for sequence modeling. By introducing input-dependent selective scanning, Mamba achieves performance comparable to Transformers on language-modeling tasks while retaining linear time complexity\cite{ref24}. Inspired by this result, a series of studies have introduced Mamba to computer vision. VMamba proposes a Cross-Scan Module that unfolds 2D feature maps into 1D sequences along four directions and uses selective scanning to establish global contextual dependencies\cite{ref25}. MambaVision further proposes a hybrid architecture: convolutional blocks extract local features in the first two stages, while hybrid modules combining Mamba SSMs and self-attention capture local details and global semantics in the later stages\cite{ref26,ref27}.

In point cloud processing, PointMamba, Point Cloud Mamba, Mamba3D, and 3DMambaIPF have begun to explore SSMs for 3D data modeling, primarily for classification, segmentation, or point cloud filtering\cite{ref28,ref29,ref30,ref31}. To the best of our knowledge, SSM designs for range-image point cloud super-resolution that jointly model local edges and global dependencies warrant further study. We therefore design LG-SSM Block to couple the local feature-extraction capability of depthwise separable convolutions with the global modeling capability of a selective state-space model within a single processing block.

\subsection{Point Cloud Denoising and Post-Processing}

Post-processing denoising is important for improving the quality of super-resolved point clouds. Traditional point cloud denoising methods mainly use filtering or learning-based approaches\cite{ref48,ref49,ref50}. Radius Outlier Removal and Statistical Outlier Removal are widely used classical filters: the former removes points with too few neighbors within a specified radius, while the latter rejects outliers based on the statistical distribution of distances to neighboring points. However, these 3D clustering methods operate directly on unstructured point clouds and cannot readily use range-image neighborhood continuity to distinguish genuine sparse details from projection artifacts.

Recent learning-based methods, such as PointCleanNet and TotalDenoising, denoise point clouds by learning point-wise displacement vectors\cite{ref49,ref50}. They typically require additional supervision and may alter genuine local geometric edges while correcting outliers. Other approaches directly refine predicted range images during post-processing, for example by using morphological operations or guided filtering to smooth anomalous regions. Our 2D median-residual denoising method belongs to this latter category. It exploits the edge-preserving property of median filtering by computing each pixel's residual relative to its local median in the range-image domain and removing pixels with anomalous residuals. This suppresses back-projection artifacts without complex 3D clustering while retaining genuine geometric edges.

\section{Method}

We propose a point cloud super-resolution method based on multi-scale spatial priors and joint local-global modeling. Given a low-resolution point cloud as input, the method ultimately produces a high-resolution point cloud obtained by inverse projection from a high-resolution distance image. The network is built from multiple LG-SSM Blocks that jointly extract multi-scale features using local convolutions and selective state-space modeling\cite{ref24,ref25,ref26,ref28,ref29,ref30,ref31}. SpatialGateFuse performs spatial-prior-guided skip-feature fusion between the encoder and decoder. Before back-projection, 2D median-residual denoising\cite{ref48,ref49,ref50} suppresses anomalous depth residuals. The overall architecture is shown in Fig.~3.

\begin{figure}[tbp]

\centering

\includegraphics[width=\linewidth]{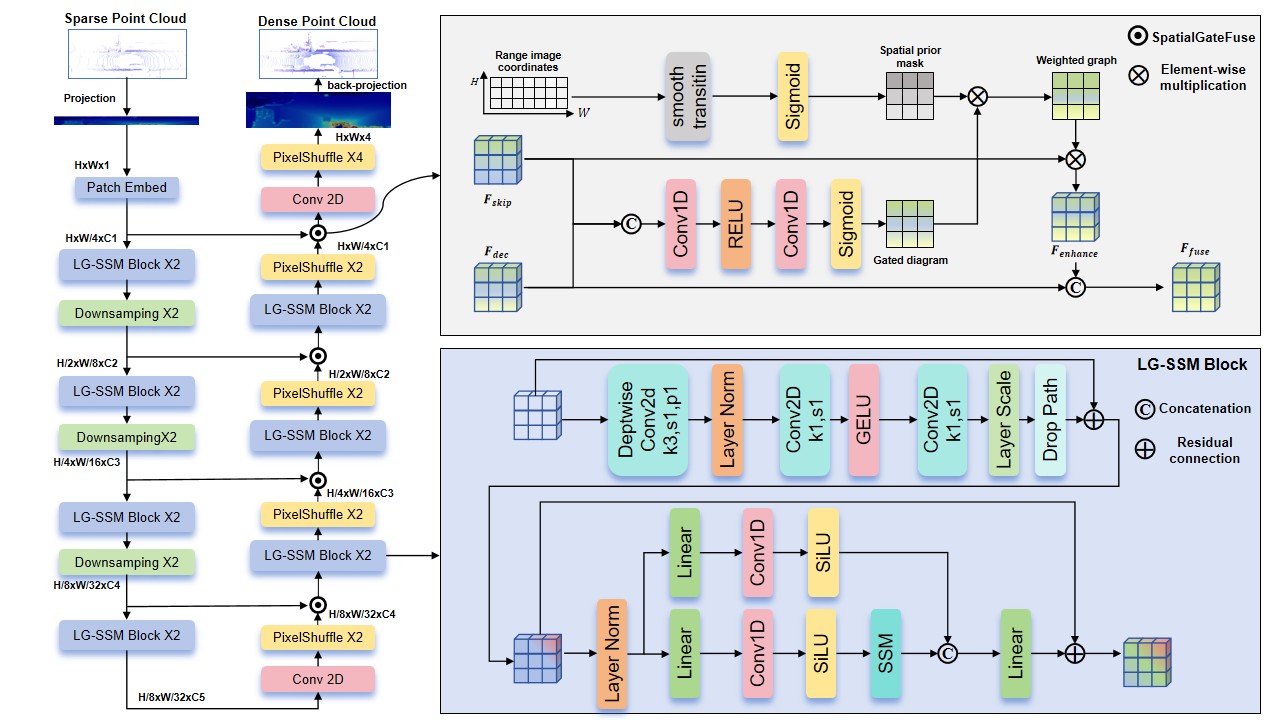}

\caption{Overall network architecture}

\label{fig:architecture}

\end{figure}

\subsection{Overall Network Framework}

The proposed method takes a low-resolution point cloud as input and obtains an initial 2D feature representation through shallow feature embedding. The encoder then progressively reduces spatial resolution and increases the channel dimension to extract multi-scale semantic features, from local edges to global structures. Unlike networks that rely solely on stacked convolutions, LG-SSM Blocks are introduced at the main encoder and decoder stages so that feature extraction captures both local geometric discontinuities and long-range continuity in range images.

During decoding, upsampling progressively restores the range-image resolution. Without reliability-aware fusion, invalid responses in shallow features from distant sparse regions can pass to subsequent decoding layers together with valid edges. We therefore insert SpatialGateFuse between shallow encoder features and deep decoder features. The module uses the vertical spatial distribution of the range image, together with the current decoder semantics, to generate content-aware gates that adaptively regulate the contribution of skip features and provide targeted structural compensation in distant sparse regions.

The high-resolution point cloud generated by backprojecting the high-resolution distance map predicted by the network output. During training, the ground-truth high-resolution range image supervises consistency between predicted and true depths. During inference and evaluation, the prediction is first denoised using 2D median-residual filtering; valid pixels are then back-projected into a 3D point cloud. Reconstruction quality is evaluated using both 2D range errors and 3D spatial-occupancy consistency. Thus, the overall framework integrates feature extraction, spatial-prior fusion, and denoising into a complete point cloud super-resolution pipeline.

\subsection{Multi-Granularity Local-Global Cooperative Perception Module}

LG-SSM Block is the core computational unit of the encoder-decoder network and extracts deep features at each encoder downsampling and decoder upsampling stage. To address the conflict between losing local high-frequency details and lacking a global long-range receptive field, each block couples local feature extraction with global reasoning in a single processing cycle.

Given an input feature map $X\in\mathbb{R}^{B\times C\times H\times W}$, where $B$ is the batch size, $C$ is the number of channels, and $H$ and $W$ are the spatial dimensions, the block performs the following operations.

The local extraction layer focuses on high-frequency spatial discontinuities in LiDAR range images, such as vehicle edges and small-obstacle contours. To accommodate the smaller feature maps and wider channels in deeper network layers, it uses depthwise separable convolutions with an inverted-bottleneck architecture to capture clear edge transitions and fine geometric textures within local neighborhoods. First, a depthwise convolution ($\operatorname{DWConv}_{3\times3}$) aggregates geometric features independently within each channel, providing an initial high-frequency edge prior: $X_{\rm dw}=\operatorname{DWConv}_{3\times3}(X)$. Layer normalization then stabilizes the feature distribution, and two pointwise (1 × 1) convolutions with GELU activations provide efficient nonlinear interactions in a higher-dimensional feature space. Specifically, the channel dimension is expanded by a factor of $r$ (set to $r=4$), passed through GELU for smooth nonlinear mapping, and then reduced to its original size. This inverted bottleneck, with narrow ends and a wider middle, helps avoid hard gradient truncation and information loss:

\[
X_{\rm local}=\operatorname{PWConv}_2(\operatorname{GELU}(\operatorname{PWConv}_1(\operatorname{LayerNorm}(X_{\rm dw}))))
\]

The high-frequency local features extracted by the inverted bottleneck are finally added to the original input through a residual connection to produce the intermediate feature $X_{\rm mid}=X_{\rm local}+X$. This representation has clearer geometric boundaries and local contour priors and serves as the input to the subsequent global reasoning layer.

After local processing, the block uses a selective state-space model to establish global dependencies across the elongated range image. It first flattens the 2D feature map with local geometric priors into a 1D sequence $x\in\mathbb{R}^{B\times L\times C}$ of length $L=H\times W$. A continuous-time state-space model maps input $x(t)$ to output $y(t)$ through a hidden state $h(t)$, with the following ordinary differential equations:

\[
h'(t)=Ah(t)+Bx(t)
\]

\[
y(t)=Ch(t)+Dx(t)
\]

To handle discrete image-feature sequences, a time-step scale parameter $\Delta$ is introduced. The zero-order hold (ZOH) rule discretizes the system to obtain the transition matrix $\bar{A}$ and input projection matrix $\bar{B}$:

\[
\bar{A}=\exp(\Delta A)
\]

\[
\bar{B}=(\Delta A)^{-1}(\exp(\Delta A)-I)\Delta B
\]

The hidden-state update and output at the current time step are then given by:

\[
h_t=\bar{A}h_{t-1}+\bar{B}x_t
\]

\[
y_t=Ch_t+Dx_t
\]

Importantly, the matrices $B$ and $C$ and the time step $\Delta$ are not fixed; they are dynamically generated from the current input token. This input-dependent selection enables the model to maintain linear time complexity $O(L)$ while adaptively bridging large sparse areas in the range image and establishing globally contextualized topological structures.

Finally, the globally processed sequence is folded back into a 2D spatial representation and combined with the backbone residual to produce the LG-SSM Block output:

\[
X_{\rm out}=\operatorname{Fold}(\operatorname{SSM}(\operatorname{Flatten}(X_{\rm mid})))+X_{\rm mid}
\]

\subsection{Spatially Gated Fusion Module}

When multi-scale features are passed from the encoder to the decoder, simply concatenating shallow and deep features along the channel dimension provides no assessment of feature reliability. Consequently, valid edges in shallow features are passed to subsequent convolutions together with invalid values and noise. SpatialGateFuse introduces content-aware gating and a vertical spatial prior to adaptively modulate skip-connection features through a residual adjustment, as described below.

Content-aware gate generation: the encoder skip feature $X_{\rm skip}$ and the current upsampled decoder feature $X_{\rm dec}$ are concatenated along the channel dimension. Successive convolution and activation layers then generate a 2D spatial attention map $G\in\mathbb{R}^{1\times H\times W}$:

\[
G=\sigma(\operatorname{Conv}_{1\times1}(\operatorname{ReLU}(\operatorname{Conv}_{1\times1}([X_{\rm skip};X_{\rm dec}]))))
\]

Here, $\sigma$ denotes the sigmoid function, and $[\cdot;\cdot]$ denotes concatenation along the channel dimension.

Spatial-prior mask generation: using the vertical coordinates of the range image, we construct a smoothly varying spatial mask $M\in\mathbb{R}^{1\times H\times1}$ whose values approach 1 in the upper (distant) region and 0 in the lower (nearby) region:

\[
M(h)=\sigma\left(\frac{h-\alpha H}{\beta H}\right)
\]

Here, $h\in[0,H-1]$ is the row index, $\alpha$ is the transition-center ratio (default 0.5), and $\beta$ controls the transition-band width (default 0.08).

Adaptive feature enhancement and fusion: a learnable gain parameter $\gamma$, initialized to zero, defines the feature amplification factor used to enhance the shallow feature:

\[
X_{\rm enhanced}=X_{\rm skip}\odot(1.0+\gamma G\odot M)
\]

The adjusted feature is then fused with the decoder feature and passed to the next layer. Zero-initializing the gain keeps the module equivalent to baseline fusion early in training, avoiding disruption of the original feature distribution before reliable gating patterns have been learned. As training proceeds, the content gate is predicted jointly from shallow and current decoder features to assess whether local responses agree with the current semantics. The vertical spatial prior concentrates additional modulation in distant sparse regions. Low gate responses therefore do not give invalid features the same enhancement as valid structures, while high responses increase the relative weight of structures consistent with decoder semantics.

\subsection{2D Median-Residual Denoising}

After the decoder's LG-SSM Blocks and PixelShuffle upsampling operations\cite{ref33}, the network produces a predicted high-resolution range image. To remove 3D point artifacts caused by interpolation-induced smoothing, we apply a 2D post-processing denoising step before back-projecting the range image, as illustrated in Fig.~4.

\begin{figure}[H]

\centering

\includegraphics[width=\linewidth]{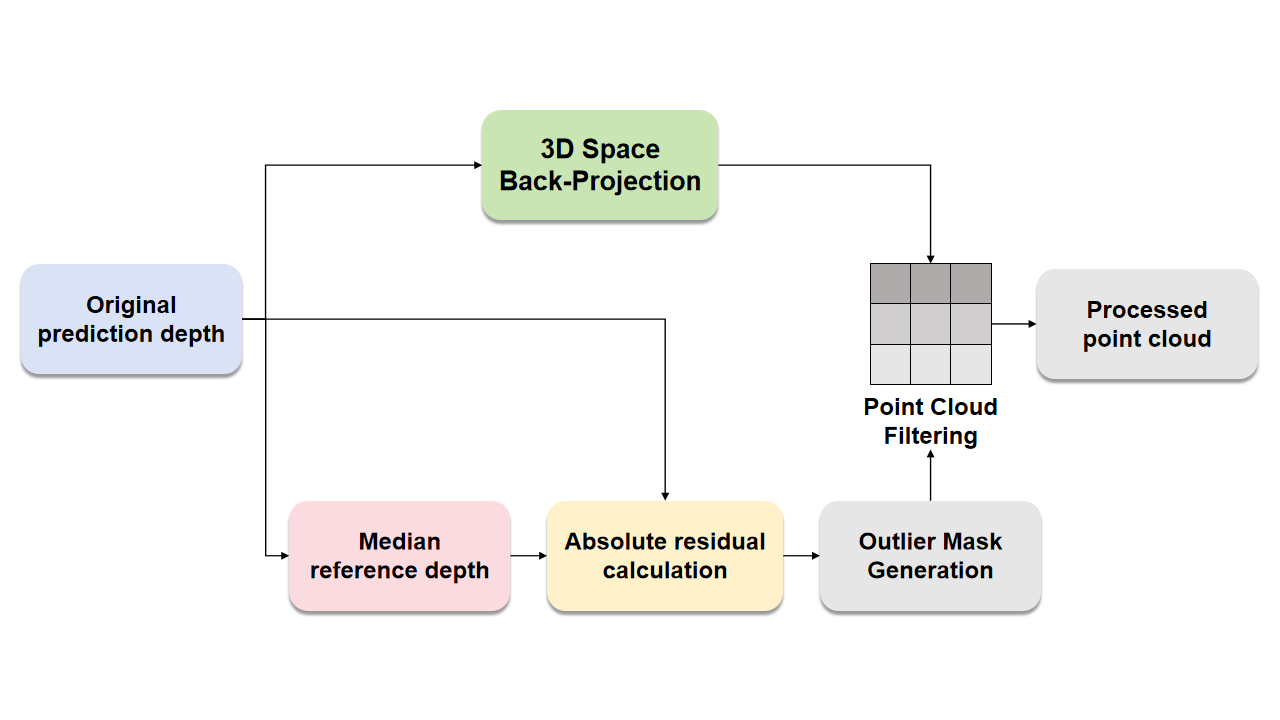}

\caption{Illustration of 2D median-residual denoising}

\label{fig:denoising}

\end{figure}

First, we estimate the locally dominant surface. A 2D median filter of size $k\times k$ (with $k=5$ in this work) is applied to the high-resolution range image to obtain a reference depth for each pixel from its neighborhood:

\[
D_{\rm med}(i,j)=\operatorname{Median}\{D_{\rm pred}(i+m,j+n)\mid m,n\in[-\lfloor k/2\rfloor,\lfloor k/2\rfloor]\}
\]

Residual-based rejection and 3D projection: the absolute residual between the predicted depth and the median reference is $R(i,j)=|D_{\rm pred}(i,j)-D_{\rm med}(i,j)|$. A dynamic outlier threshold $\tau$ is used. When each high-resolution pixel is back-projected using the camera intrinsics to form a dense 3D point cloud, pixels with $R(i,j)>\tau$ are treated as interpolation artifacts and excluded; all other pixels are projected normally.

Median filtering is effective at suppressing noise while preserving edges. At boundaries between foreground objects and distant backgrounds, incorrectly smoothed transition depths may have large residuals and are removed by this rule. The method suppresses floating outliers and fragmented artifacts through residual checks in the 2D image domain, without expensive clustering in high-dimensional 3D space.

\section{Experiments}

\subsection{Experimental Setup}

Datasets. We evaluate the method on two widely used autonomous-driving LiDAR datasets\cite{ref51,ref52}.

KITTI-360\cite{ref51}: captured using a 64-beam Velodyne HDL-64E LiDAR, this dataset contains approximately 20,000 training samples and 2,500 validation samples. The low-resolution input is obtained by downsampling from 64 to 16 beams, and the target is the original 64-beam resolution.

nuScenes\cite{ref52}: captured using a 32-beam Velodyne HDL-32E LiDAR, this dataset is processed from an 8-beam input to a 32-beam target resolution. 

Evaluation metrics. We evaluate reconstruction quality using metrics at both the 2D and 3D levels:

2D metric: mean absolute error (MAE), which measures the per-pixel absolute difference between the predicted and ground-truth range images.

Chamfer Distance (CD)\cite{ref53}: the symmetric nearest-neighbor distance between the predicted and ground-truth point clouds, reflecting overall 3D reconstruction accuracy.

Voxel-based metric: intersection over union (IoU), calculated on a voxel grid with a resolution of 0.1 m, to measure spatial-occupancy agreement between point clouds.

\subsection{Comparative Experiments}

\subsubsection{Qualitative Results}

To visually assess recovery of scene structure and local detail, Figs.~5 and 6 compare sparse inputs, ground-truth high-resolution point clouds, our method, TULIP, SRFormer, and SRMamba. The qualitative evaluation focuses on two issues in range-image super-resolution: excessive smoothing around depth transitions, which can merge neighboring objects or object-background contours and distort the original discrete point distribution; and isolated spike artifacts caused by depth prediction errors after back-projection. Red boxes and insets highlight structural preservation around depth discontinuities.

\begin{figure}[tbp]

\centering

\includegraphics[width=\linewidth]{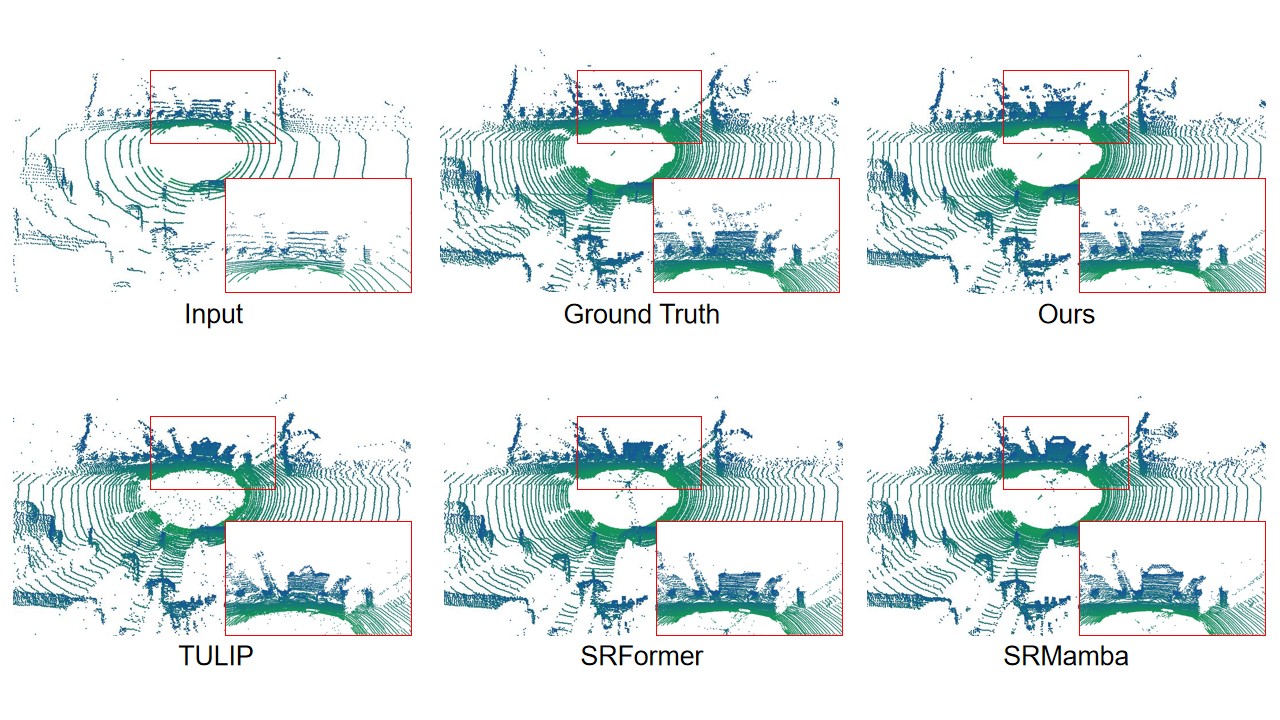}

\caption{Comparison of point cloud over-smoothing across methods}

\label{fig:qualitative-smoothing}

\end{figure}

As shown in Fig.~5, sparse inputs lack continuous depth observations around point cloud contours and their boundaries with the background. If neighboring depths are over-smoothed during super-resolution, depth layers that should remain separate can merge into thick bands, causing local detail loss and contour adhesion. In the red-boxed regions, some competing methods increase point density but over-smooth the original discrete structure, leading to misinterpretation of point cloud features. By contrast, our reconstruction is closer to the ground truth overall: it recovers the main structures and their upper contours while maintaining separation between road scan lines and foreground objects. This indicates that our method supplements sparse regions while constraining feature propagation near boundaries, preserving depth discontinuities and reducing structural mixing caused by smoothing.

Figure 6 further compares detail recovery and outlier suppression in complex sparse scenes. Although the competing methods reconstruct the main scene contours, spike artifacts and local adhesion remain visible around vehicles, along vertical object boundaries, and at depth discontinuities. The point clouds generated by our method show greater consistency with the ground truth in their main structures, boundary trajectories, and local point distributions. After 2D median-residual post-processing, outliers floating beyond the main surfaces are further suppressed, while valid structures such as vehicles and road edges are preserved. Thus, our method not only increases point-cloud density but also improves spatial continuity and boundary delineation in the reconstruction.

\begin{figure}[!htbp]
\centering
\includegraphics[width=\linewidth]{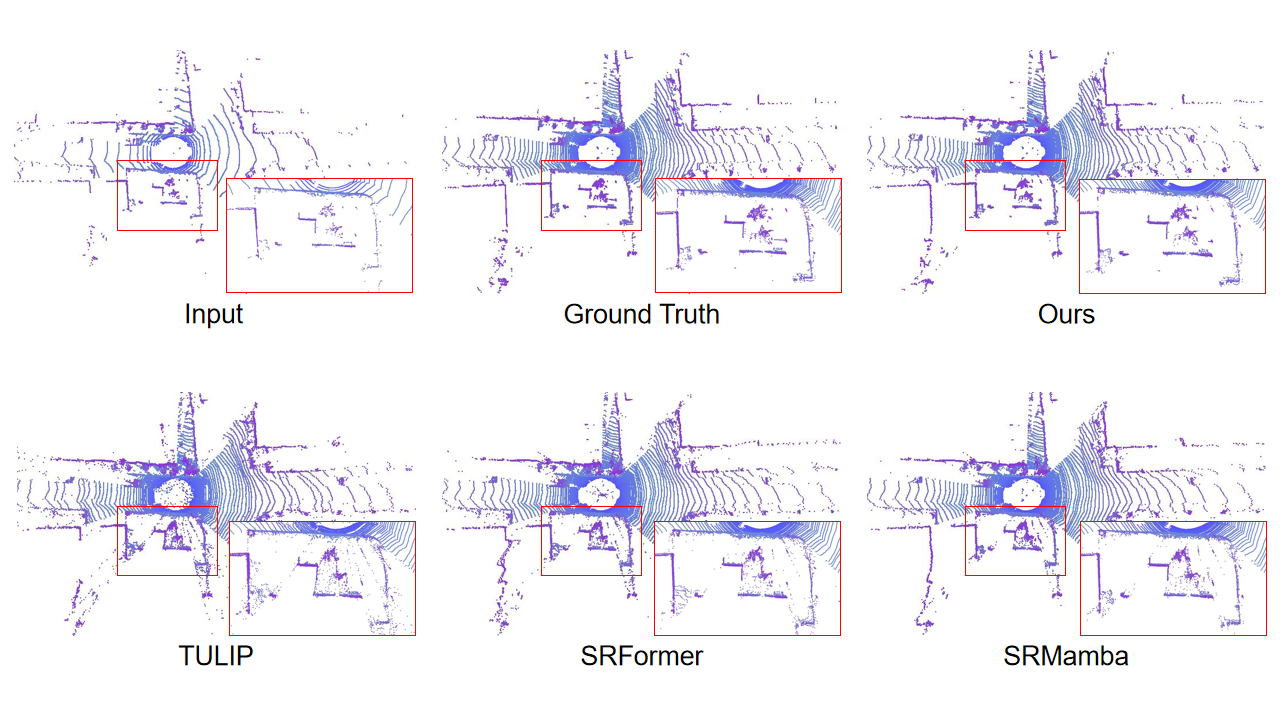}
\caption{Detail recovery and outlier suppression in complex sparse scenes}
\label{fig:qualitative-denoising}
\end{figure}

\subsubsection{Quantitative Results}

Our method achieves an IoU of 0.4703 on KITTI-360 and 0.3593 on nuScenes. These results show that the proposed pipeline performs range-image-based point cloud super-resolution across datasets with different beam configurations. The specific test indicators are provided in Tables 1 and 2.

\begin{table}[tbp]
\centering
\caption{Quantitative comparison on the KITTI-360 dataset}
\label{tab:kitti-results}
\begin{tabular}{llccc}
\hline
Methods & Dataset & MAE & CD & IoU \\
\hline
VIT & KITTI-360 & 0.0056 & 0.1192 & 0.4011 \\
ConvNext & KITTI-360 & 0.0051 & 0.1215 & 0.4041 \\
SRFormer & KITTI-360 & 0.0065 & 0.1374 & 0.4110 \\
TULIP & KITTI-360 & 0.0051 & 0.1241 & 0.4152 \\
EchoSR & KITTI-360 & 0.0102 & 0.1260 & 0.4343 \\
SRMamba & KITTI-360 & 0.0044 & 0.1031 & 0.4389 \\
Ours & KITTI-360 & 0.0043 & 0.0563 & 0.4703 \\
\hline
\end{tabular}
\end{table}

\begin{table}[tbp]
\centering
\caption{Quantitative comparison on the nuScenes dataset}
\label{tab:nuscenes-results}
\begin{tabular}{llccc}
\hline
Methods & Dataset & MAE & CD & IoU \\
\hline
SRFormer & nuScenes & 0.0297 & 1.3405 & 0.2906 \\
TULIP & nuScenes & 0.0293 & 1.0502 & 0.3048 \\
SRMamba & nuScenes & 0.0287 & 1.0196 & 0.317 \\
Ours & nuScenes & 0.0302 & 1.0065 & 0.3593 \\
\hline
\end{tabular}
\end{table}

\subsection{Ablation Study}

To evaluate the effectiveness of the proposed components, we progressively introduce and combine LG-SSM Block, SpatialGateFuse (SGF), and 2D median-residual post-processing on KITTI-360.

\begin{table}[H]

\centering

\caption{Ablation results for the core modules}

\begin{tabular}{ccccccc}

\hline

Basic [2, 2, 2, 2] & LG & SGF & Post-processing & MAE & CD & IoU \\

\ding{51} &  &  &  & 0.0044 & 0.1031 & 0.4389 \\

\ding{51} & \ding{51} &  &  & 0.0045 & 0.0879 & 0.4430 \\

\ding{51} &  & \ding{51} &  & 0.0044 & 0.0877 & 0.4454 \\

\ding{51} & \ding{51} & \ding{51} & \ding{51} & 0.0043 & 0.0563 & 0.4703 \\

\hline

\end{tabular}

\label{tab:ablation}

\end{table}

\section{Conclusion}

We addressed three challenges in range-image-based point cloud super-resolution: inadequate feature transfer under nonuniform sampling, difficulty balancing local details with global dependencies, and artifacts introduced during 2D-to-3D back-projection. We proposed a method based on multi-scale spatial priors and joint local-global modeling. SpatialGateFuse enhances useful features in distant sparse regions through spatial gating. LG-SSM Block combines local convolution with selective state-space modeling to preserve high-frequency edges and long-range structural dependencies. 2D median-residual denoising uses an edge-preserving residual test in the range-image domain to suppress spike artifacts. Experiments on KITTI-360 and nuScenes demonstrate consistent improvements, showing that the three modules work together to enhance point cloud super-resolution quality.

\end{document}